\documentclass[journal]{IEEEtai}

\usepackage[colorlinks,urlcolor=blue,linkcolor=blue,citecolor=blue]{hyperref}

\usepackage{color,array}
\usepackage[table]{xcolor} 
\usepackage{graphicx}

\usepackage{multirow}

\usepackage{comment}
\usepackage{amsmath}
\usepackage{amssymb}
\begin{document}

\title{SMTrack: End-to-End Trained Spiking Neural Networks for Multi-Object Tracking in RGB Videos}


\author{Pengzhi Zhong, Xinzhe Wang, Dan Zeng, Qihua Zhou, Feixiang He, and Shuiwang Li$^*$\thanks{* Corresponding author.}
\thanks{Pengzhi Zhong and Shuiwang Li are with the College of Computer Science and Engineering, Guilin University of Technology, China, 541006 and Guangxi Key Laboratory of Embedded Technology and Intelligent System, Guilin University of Technology, China, 541004 (e-mail: zhongpengzhi@glut.edu.cn; lishuiwang0721@163.com).}
\thanks{Xinzhe Wang and Qihua Zhou are with the College of CompuerScience and Software Engineering, Shenzhen University,Shenzhen 518060, China (e-mail: 2500101066@mails.szu.edu.cn; qihuazhou@szu.edu.cn).}
\thanks{Dan Zeng is with the School of Artificial Intelligence, Sun Yat-sen University, Guangzhou 510275, China (e-mail: zengd8@mail.sysu.edu.cn).}
\thanks{Feixiang He is with the School of Electronic Information, Central South University, Changsha 410083, China (e-mail: feixiang.he@csu.edu.cn).}
}

\markboth{}
{}

\maketitle
\begin{abstract}
Brain-inspired Spiking Neural Networks (SNNs) exhibit significant potential for low-power computation, yet their application in visual tasks remains largely confined to image classification, object detection, and event-based tracking. In contrast, real-world vision systems still widely use conventional RGB video streams, where the potential of directly-trained SNNs for complex temporal tasks such as multi-object tracking (MOT) remains underexplored. To address this challenge, we propose SMTrack—the first directly trained deep SNN framework for end-to-end multi-object tracking on standard RGB videos. SMTrack introduces an adaptive and scale-aware Normalized Wasserstein Distance loss (Asa-NWDLoss) to improve detection and localization performance under varying object scales and densities. Specifically, the method computes the average object size within each training batch and dynamically adjusts the normalization factor, thereby enhancing sensitivity to small objects. For the association stage, we incorporate the TrackTrack identity module to maintain robust and consistent object trajectories. Extensive evaluations on BEE24, MOT17, MOT20, and DanceTrack show that SMTrack achieves performance on par with leading ANN-based MOT methods, advancing robust and accurate SNN-based tracking in complex scenarios.

\end{abstract}



\section{Introduction}

\IEEEPARstart
Multi-object tracking (MOT) plays a vital role in numerous computer vision applications, such as intelligent surveillance, autonomous navigation, and smart transportation \cite{oh2011large,sportsmot,Deepdriving,zhang2019robust,xiang2015learning}. With the increasing deployment of MOT systems on edge devices like drones and mobile robots, real-time performance and energy efficiency have become critical requirements \cite{ByteTrack}. However, conventional MOT algorithms based on artificial neural networks (ANNs) face significant challenges in terms of latency and power consumption \cite{tu2021resource,li2023multiobject}. In contrast, spiking neural networks (SNNs)—inspired by biological neural mechanisms—have shown great potential for low-power neuromorphic systems due to their sparse and event-driven computation paradigm \cite{maass1997networks,roy2019towards,merolla2014million,poon2011neuromorphic}.

Currently, most research applying SNNs to visual tasks has primarily focused on image classification \cite{deng2022temporal,EM-ResNet,guo2023rmp}, object detection \cite{Spiking-yolo,Meta-SpikeFormer,EMS-YOLO,SpikeYOLO}, and event-based object tracking \cite{mitrokhin2018event,STNet,zheng2022spike,qu2024spike} using dynamic vision sensors (DVS). Although the sparse and asynchronous data stream produced by DVS aligns well with the processing characteristics of SNNs, RGB image sequences remain the dominant input modality in real-world visual systems. Consequently, developing real-time, spike-driven MOT algorithms based on conventional RGB video inputs is of significant importance for both future research and practical deployment. While ANN-to-SNN conversion methods have made notable progress in recent years, they typically rely on long temporal windows, which limits their suitability for MOT tasks that demand both efficiency and accuracy \cite{NeuroSORT}. For example, Spiking-YOLO \cite{Spiking-yolo} requires at least 3,500 timesteps to approximate the original ANN’s performance, while Spike Calib \cite{SpikeCalib} reduces the timestep count to a few hundred but still heavily depends on the quality of a pretrained ANN. In contrast, directly trained SNNs based on surrogate gradient methods offer a more promising alternative, achieving comparable or superior performance with significantly fewer timesteps \cite{MSResNet,EM-ResNet}. To advance the direct training of SNNs for object detection tasks, recent works such as EMS-YOLO \cite{EMS-YOLO} and SpikeYOLO \cite{SpikeYOLO}, have proposed energy-efficient and high-performance SNN architectures tailored for object detection.
SpikeYOLO integrates the macro-architecture of YOLOv8 with the micro-level Meta-SpikeFormer modules and introduces the Integer Leaky Integrate-and-Fire (I-LIF) neuron, unifying integer training with spike-based inference. It has achieved detection accuracy comparable to mainstream artificial neural networks (ANNs) on COCO \cite{COCO} and Gen1 \cite{Gen1} datasets. While these advancements demonstrate the potential of directly trained SNNs for object detection, the development of a comprehensive, real-time, spike-driven Multi-Object Tracking (MOT) framework based on conventional RGB video inputs remains a significant challenge. MOT scenarios involve substantial scale variations, encompassing large, nearby objects as well as small, distant ones, which demand detectors with strong multi-scale modeling capabilities. To address this, we propose SMTrack, the first directly-trained deep spiking neural network (SNN) framework for end-to-end multi-object tracking from standard RGB video streams. Regarding the challenge of scale variation, the conventional Intersection over Union (IoU) loss \cite{iou}, though widely used, is overly sensitive to localization shifts for small objects. Although the NWDloss \cite{NWDloss} alleviates this issue, its normalization term depends on a fixed scaling constant, typically set based on the average object size across the entire dataset. When the dataset contains objects with uneven scale distribution, this fixed factor may fail to adapt to multi-object tracking scenarios, thereby degrading overall detection performance. To overcome this, we propose an adaptive and scale-aware NWDloss (Asa-NWDLoss), which dynamically adjusts the normalization factor based on the object size distribution within each batch. This strategy enables more robust and accurate bounding box regression under multi-scale and multi-density conditions. Furthermore, SMTrack integrates a YOLOX-inspired \cite{yolox} decoupled spiking detection head, which preserves the computational advantages of SNN while aligning its architecture with the YOLOX detector commonly used in modern MOT. This design enables more direct and fair comparisons with existing ANN-based MOT approaches. For the identity association stage, we integrate the TrackTrack \cite{TrackTrack} framework, which brings state-aware, two-stage association to the SNN pipeline. TrackTrack exhibits strong robustness in handling occlusions and demonstrates excellent identity preservation capability, ultimately achieving more stable and reliable MOT performance. The main contributions of this paper are as follows:

\begin{itemize}

\item We propose SMTrack, the first directly-trained deep spiking neural network (SNN) framework for end-to-end multi-object tracking from standard RGB video streams.

\item  We introduce an an adaptive and scale-aware Normalized Wasserstein Distance loss (Asa-NWDLoss), in which the normalization factor is dynamically adjusted based on the average object size within each batch. This improves localization stability for multi-scale and imbalanced targets.

\item Extensive experiments on four benchmarks—BEE24, MOT17, MOT20, and DanceTrack—demonstrate that SMTrack, with just four time steps, achieves state-of-the-art performance among SNN-based trackers and competitive accuracy compared with ANN-based counterparts.

\end{itemize}

\begin{figure*}[t]
	\centering
	\includegraphics[width=1.0\textwidth]{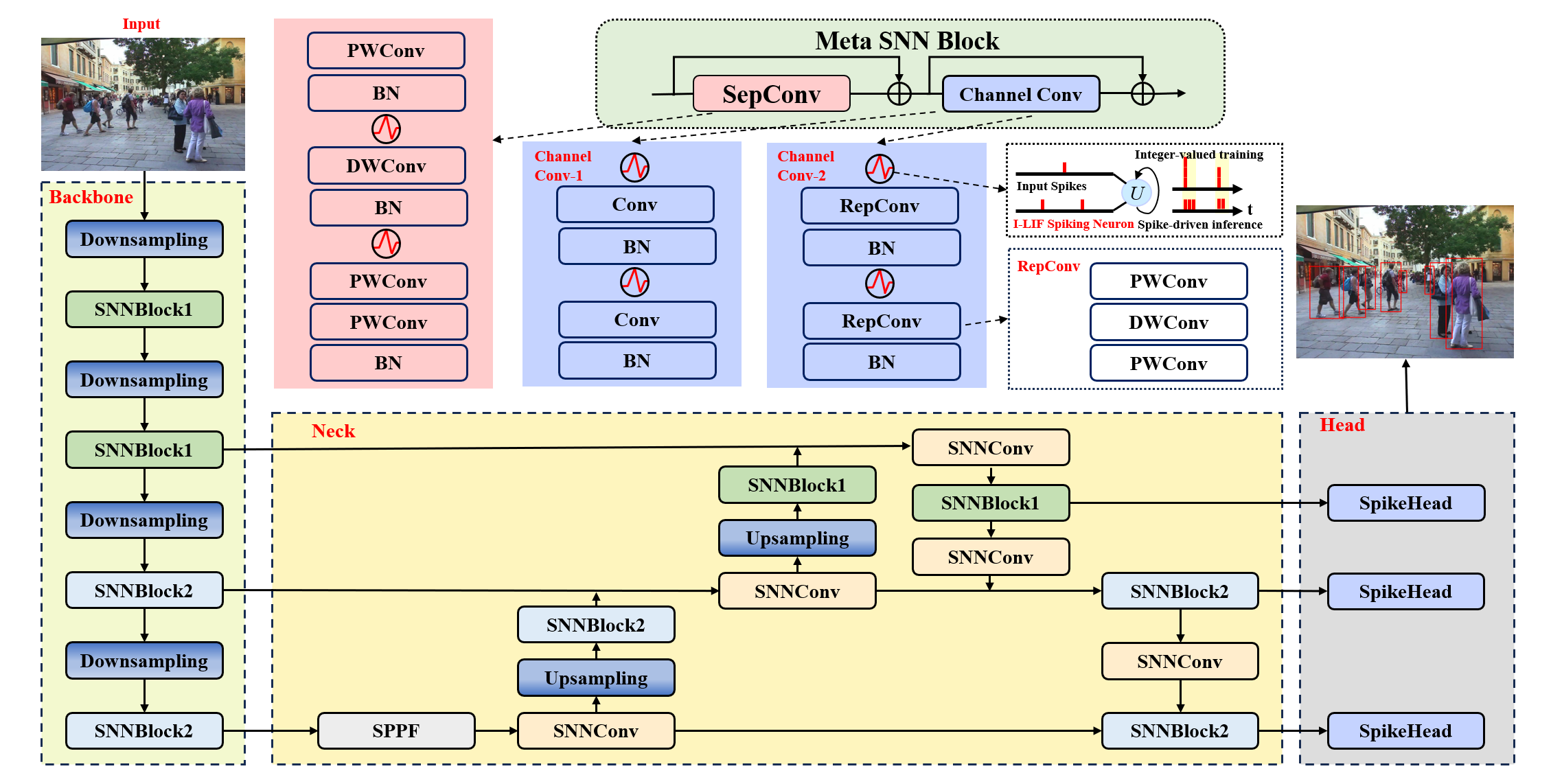}	\vspace{-0.15in}
	\caption{\small The overall framework of SpikeYOLOX, which inherited from SpikeYOLO. The difference lies in the detection head.
}
	\label{fig:SpikeYOLOX}	
\end{figure*}

\section{Related Works}

\subsection{Spiking Neural Networks for Object Detection}

Spiking Neural Networks (SNNs) have attracted increasing attention in the field of computer vision in recent years due to their event-driven computation paradigm and high energy efficiency \cite{deng2022temporal,EM-ResNet,guo2023rmp,liu2022ultralow}, Early deep SNN architectures for object detection primarily relied on conversion from Artificial Neural Networks (ANNs), For instance, Spiking-YOLO \cite{Spiking-yolo} adopts an ANN-to-SNN conversion strategy and introduces channel normalization and signed neurons to alleviate performance degradation in deep networks, achieving competitive detection accuracy. However, it requires at least 3500 timesteps to approach the performance of its original ANN counterpart, limiting its applicability in real-time and energy-constrained scenarios. To address this, Spike-Calib \cite{SpikeCalib} proposes a spike calibration (SpiCalib) technique to mitigate the distortion of output distributions caused by discrete spikes, reducing the required timesteps to a few hundred—though it still depends on a pre-trained ANN. SEENNs \cite{Seenn} further introduce a dynamic timestep adjustment mechanism that adapts the number of timesteps based on the characteristics of each input sample, maintaining high accuracy while significantly lowering the average inference latency. Although these methods mitigate temporal delay to some extent, they still rely on extended inference windows to approximate the continuous behavior of ANNs, making them less suitable for low-latency, low-power applications. In contrast, fully spike-based SNN architectures trained via end-to-end methods \cite{EM-ResNet,MSResNet} are considered more promising, offering a better trade-off between accuracy and efficiency. For example, EMS-YOLO \cite{EMS-YOLO} is the first to build a directly trained SNN detector using LIF neurons, achieving effective inference within just 4 timesteps, thereby demonstrating the feasibility of fully spiking approaches in object detection. Meta-SpikeFormer \cite{Meta-SpikeFormer} further incorporates pre-training and fine-tuning mechanisms to enhance task adaptability. Building on this, SpikeYOLO \cite{SpikeYOLO} simplifies the YOLO \cite{YOLO} architecture and integrates lightweight meta SNN blocks. Combined with integer-based training and virtual timestep expansion during inference, it effectively reduces quantization error and significantly narrows the performance gap between SNNs and ANNs in object detection tasks. Although directly trained SNNs typically underperform compared to ANN-SNN conversion-based structures in terms of accuracy, they offer greater design flexibility and require significantly fewer timesteps. In this work, we adopt a directly trained SNN for the detection module in multi-object tracking, aiming to leverage its advantages in architectural adaptability and computational efficiency.

\subsection{Multi-object tracking}

Multi-object tracking (MOT) has long been dominated by the tracking-by-detection (TBD) paradigm, which decomposes the tracking task into two stages: object detection and cross-frame association \cite{bewley2016simple,StrongSORT,he2021learnable,QDTrack}. TBD approaches typically construct an association matrix to quantify the similarity between historical trajectories and current detections, and use the Hungarian algorithm \cite{Hungarian} to perform optimal matching. The association matrix is commonly built using either motion information \cite{Mat,khurana2021detecting,Motiontrack} or appearance features \cite{wojke2017simple,kim2021discriminative,QDTrack}. However, both types of models have limitations: motion-based models are sensitive to occlusion and non-linear movements, while appearance-based models can be disrupted by visually similar targets or appearance changes. To enhance robustness, hybrid approaches fuse motion and appearance cues \cite{geiger2013vision,sportsmot,Bot-sort}. For instance, TrackFlow \cite{Trackflow} introduces a probabilistic framework to handle association uncertainty. However, TrackFlow's reliance on synthetic training data limits its generalizability.
In recent years, the joint detection and tracking (JDT) paradigm has emerged, enabling end-to-end learning of detection and appearance embedding via a shared backbone network \cite{JDE,Mots}. Early JDT methods such as JDE \cite{JDE} suffered from performance drops due to the high training complexity of the re-identification branch. To address this, FairMOT \cite{Fairmot} introduced a balanced architecture that jointly optimizes detection and re-identification tasks, significantly surpassing traditional TBD methods in accuracy. Meanwhile, Transformer architectures and attention mechanisms have been applied to appearance modeling \cite{Transtrack,Trackformer}, and several studies have further incorporated scene context to enhance feature robustness \cite{li2023inference,Qdtrack1}.
During the data association stage, most existing methods rely on computing appearance embedding similarities between detections and trajectories. However, in scenarios involving occlusions or targets with similar appearances, the reliability of such methods remains limited \cite{dancetrack}. To improve robustness, researchers have begun incorporating low-level visual cues. Optical flow-based methods have evolved from FlowNet \cite{Flownet}, with its end-to-end learning design, to RAFT \cite{Raft}, which adopts a recurrent refinement architecture that better handles small objects and fast-moving targets. Geometric correspondence models such as MatchFlow \cite{matchflow} construct a 4D correlation volume to unify optical flow estimation with cross-view matching. GeneralTrack takes a low-level vision approach, aiming to establish pixel-to-instance associations to enhance tracking accuracy in complex dynamic scenes. Recently, TOPICTrack \cite{TOPICTrack} introduces a parallel association paradigm that jointly leverages motion and appearance features. By adaptively selecting the most informative modality based on motion intensity, it improves tracking performance in scenarios with complex motion and diverse environments. TrackTrack \cite{TrackTrack} has improved the accuracy and robustness of multi-object tracking by introducing two key strategies: Track-Perspective Association (TPA) and Track-Aware Initialization (TAI). These strategies enable more precise association between detections and tracks while effectively suppressing redundant track generation. In this paper, we employ TrackTrack as the association module in SMTrack to fully leverage its strengths in detection assignment and track management.

\begin{figure}[t]
	\centering
	\includegraphics[width=1.0\linewidth]{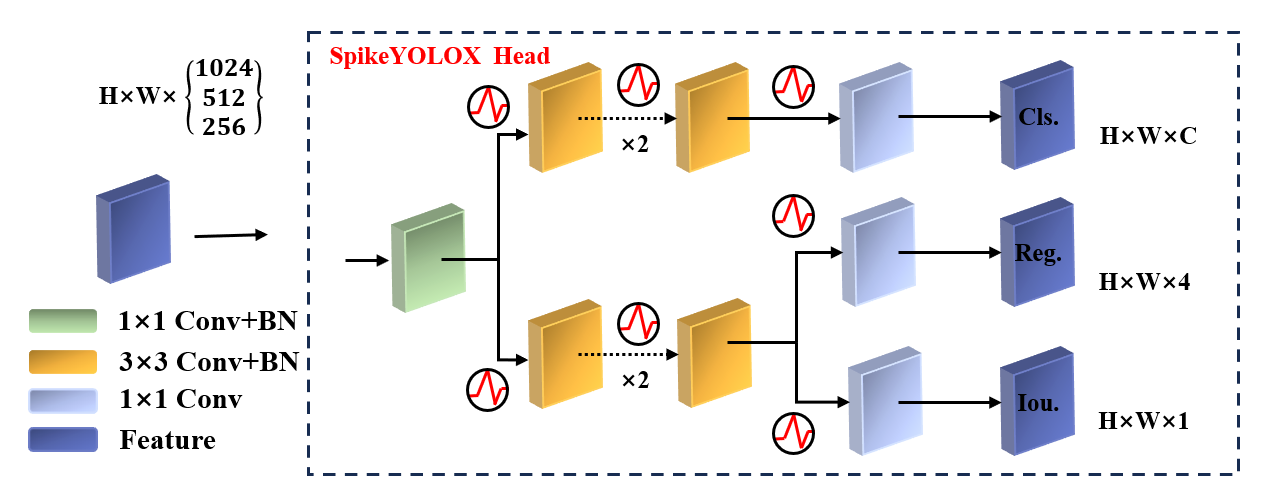}
	\vspace{-0.1in}
	\caption{\small Illustrates the SpikeYOLOX head, a spiking detection head inspired by YOLOX, where the spiking neurons are implemented using the I-LIF.}
	\label{fig:SpikeYOLOXHead}
\end{figure}

\section{Methods}
In this section, we present SMTrack, the first directly trained spiking neural network (SNN) framework designed for end-to-end multi-object tracking (MOT) from conventional RGB video streams. As illustrated in Figure~\ref{fig:SpikeYOLOX}, the proposed framework consists of two main components: (1) the SpikeYOLOX detector, an improved SNN-based object detector adapted from SpikeYOLO, and (2) an identity association module based on the TrackTrack~\cite{TrackTrack} algorithm. We first provide a detailed description of the detection module architecture, including the decoupled spiking prediction head and the proposed Asa-NWDLoss, which enhances localization stability under multi-scale object scenarios. Finally, we present the identity association pipeline built upon TrackTrack \cite{TrackTrack}, enabling robust trajectory association and identity preservation in challenging environments.
\subsection{SpikeYOLOX} To ensure direct and fair comparisons with mainstream ANN-based MOT methods, we construct SpikeYOLOX upon the widely adopted YOLOX framework, retaining its overall architecture so that our model remains structurally consistent with existing ANN-MOT algorithms. Meanwhile, SpikeYOLO has already demonstrated detection accuracy on static image tasks comparable to that of mainstream artificial neural networks (ANNs), showcasing its strong feature extraction capabilities. Based on this, we integrate the channel-mixing modules, designed with spiking neural network (SNN) meta-architecture from SpikeYOLO, into the backbone and neck of SpikeYOLOX, while employing a YOLOX-style decoupled spiking detection head. This design preserves the computational characteristics of SNNs while maintaining structural consistency with commonly used ANN-based MOT systems. To support efficient training and inference, SpikeYOLOX employs Integer Leaky Integrate-and-Fire (I-LIF) neurons, which activate discrete integer-valued membrane potentials during training while maintaining spike-driven dynamics during inference by simulating virtual timesteps. In our SpikeYOLOX, the maximum emitted integer value of I-LIF is set to $D=4$, balancing accuracy and efficiency.

\subsubsection{Channel Mixing Module} The meta-design of SNNs integrates modules based on both convolutional neural networks (CNNs) and Transformer architectures. Each meta-block consists of a token mixer and a channel mixer. CNN-based SNN modules use spike-driven convolution as the token mixer, while Transformer-based modules adopt spike-driven self-attention mechanisms. In SpikeYOLO \cite{SpikeYOLO}, the channel mixer was specifically optimized for object detection tasks. Specifically, as illustrated in Figure \ref{fig:SpikeYOLOX}, SNN-Block-1 and SNN-Block-2 are respectively designed to extract low-level and high-level feature representations from input images.
The meta SNN block is defined as:
{\small
\begin{equation}
Y' = Y + \mathrm{SepConv}(Y)
\end{equation}
\begin{equation}
Y'' = Y' + \mathrm{ChConv}(Y')
\end{equation}
}
where $Y \in \mathbb{R}^{T \times C \times H \times W}$ denotes the input tensor of the layer. The $\mathrm{SepConv}(\cdot)$ module adopts the design of MobileNetV2 and employs a depthwise separable convolution with a large kernel size to extract global spatial features. It is formulated as:
{\small
\begin{equation}
    \mathrm{Z}(Y)  =  \mathrm{DWConv}_1\left( \mathrm{ILIF}\left( \mathrm{PWConv}_1\left( \mathrm{ILIF}(Y) \right) \right) \right)
\end{equation}
\begin{equation}
\mathrm{SepConv}(Y) = \mathrm{ILIF}\left(\mathrm{DWConv}_2\left( \mathrm{PWConv}_2\left( \mathrm{Z}\right) \right) \right) 
\end{equation}
}

$\mathrm{PWConv}_i(\cdot)$ and $\mathrm{DWConv}_i(\cdot)$ refer to point-wise and depth-wise convolutions, respectively \cite{Xception}, and $\mathrm{ILIF}(\cdot)$ denotes the I-LIF spiking neuron activation function. The $\mathrm{ChConv}(\cdot)$ module serves as a channel-wise feature fusion mechanism. We adopt different channel mixing strategies in the lower and higher layers:

For SNN-Block-1, the operation is formulated as:
{\small
\begin{equation}
\mathrm{ChConv}_1(Y') = \mathrm{Conv}\left( \mathrm{ILIF}\left( \mathrm{Conv}\left( \mathrm{ILIF}(Y') \right) \right) \right)
\end{equation}
}
Here, $\mathrm{Conv}(\cdot)$ denotes a standard convolutional layer, with an expansion factor set to $r=4$ to strengthen cross-channel feature interactions. In contrast, SNN-Block-2, designed for higher-level feature representation—introduces a lightweight structure based on re-parameterized convolution to reduce model complexity. It is defined as:
{\small
\begin{equation}
\mathrm{ChConv}_2(Y') = \mathrm{RepConv}\left( \mathrm{ILIF}\left( \mathrm{RepConv}\left( \mathrm{ILIF}(Y') \right) \right) \right)
\end{equation}
\begin{equation}
\mathrm{RepConv}(Y') = \mathrm{PWConv}_2 \left( \mathrm{DWConv}_1 \left( \mathrm{PWConv}_1(Y') \right) \right)
\end{equation}
}

This re-parameterized block employs a $3 \times 3$ convolutional kernel and can be equivalently transformed into a standard convolution during the inference phase, significantly reducing computational overhead without sacrificing representational capacity.

\begin{table*}[h]
    \centering
        \caption{Comparison with state-of-the-art methods on the DanceTrack test set. Metrics marked with $\uparrow$ indicate that higher values are better, and vice versa. Bold and blue values represent the best and second-best results in each column, respectively.}
    \renewcommand{\arraystretch}{1.3} 

    \begin{tabular}{lccccccccc}
        \hline
        \textbf{Tracker} & \textbf{Venue} & \textbf{HOTA $\uparrow$} & \textbf{MOTA $\uparrow$} & \textbf{IDF1 $\uparrow$} & \textbf{AssA $\uparrow$} & \textbf{AssR $\uparrow$} & \textbf{DetA $\uparrow$} & \textbf{IDs $\downarrow$} & \textbf{Frag $\downarrow$} \\
        \hline
        ByteTrack & ECCV22 & 43.36 & 86.50 & 48.99 & 27.56 & 34.43 & 68.49 &  958  & 3885 \\
        OC-SORT & CVPR23 & 48.00 & 87.09 & 47.82 & 30.22 & 35.30 & 76.58 & 995 & 4247 \\
        
        Deep OC-SORT & ICIP23 & 49.53 & \textbf{90.01} & 48.99 & 31.52 & 36.76 & 78.21 & 1027 & \textbf{\textcolor{blue}{2476}} \\

        AdapTrack & ICIP24 & 49.92 & 83.98 & 48.53 & 34.27 & 37.98 & 73.45 & 1775 & \textbf{2403} \\
        CMTrack & ICIP24 & 54.61 & 89.50 & 55.45 & 38.20 & 44.96 & 78.38 & 898 & 3676 \\
        HybridSORT & AAAI24 & 56.10 & 89.40 & 57.16 & 40.33 & 46.20 & 78.37 & 923 & 3646 \\
        TOPICTrack & TIP25 & 49.76 & \textbf{\textcolor{blue}{89.92}} & 47.75 & 31.63 & 35.83 & \textbf{\textcolor{blue}{78.69}} & 1207 & 2500 \\
        TrackTrack & CVPR25 & \textbf{\textcolor{blue}{56.54}} & 89.57 & \textbf{\textcolor{blue}{57.21}} & \textbf{\textcolor{blue}{40.93}} & \textbf{\textcolor{blue}{46.50}} & 78.44 & \textbf{\textcolor{blue}{922}} & 3448 \\
        \hline
         \rowcolor{gray!20} %
        SMTrack & Ours & \textbf{57.76} & 89.62 & \textbf{58.47} & \textbf{42.62} & \textbf{47.91} & \textbf{78.70} & \textbf{857} & 3506 \\
        \hline
    \end{tabular}

    \label{tab:DanceTrack}
\end{table*}

\begin{table*}[h]
    \centering
    \caption{Comparison with state-of-the-art methods on the BEE24 test set. Metrics marked with $\uparrow$ indicate that higher values are better, and vice versa. Bold and blue values represent the best and second-best results in each column, respectively.}
    \renewcommand{\arraystretch}{1.3} 
    \begin{tabular}{lccccccccc}
        \hline
        \textbf{Tracker} & \textbf{Venue} & \textbf{HOTA $\uparrow$} & \textbf{MOTA $\uparrow$} & \textbf{IDF1 $\uparrow$} & \textbf{AssA $\uparrow$} & \textbf{AssR $\uparrow$} & \textbf{DetA $\uparrow$} & \textbf{IDs $\downarrow$} & \textbf{Frag $\downarrow$} \\
        \hline
         ByteTrack & ECCV22 & 46.62 & 65.31 & 59.67 & 41.10 & 60.50 & 53.36 & 1166 & 2425 \\
        OC-SORT & CVPR23 &  50.83 & 72.08 & 64.72 & \textbf{\textcolor{blue}{45.88}} &  62.60  & \textbf{\textcolor{blue}{56.70}} & 1397 & 2578 \\
        Deep OC-SORT & ICIP23 & 49.22 & 67.76 & 62.81 & 44.37 & 61.34 & 54.94 & 1383 & 2031 \\

        AdapTrack & TIP24 & 47.87 &	60.77 &  61.07&	44.18	&\textbf{\textcolor{blue}{63.48}}	&52.25	&1451	&2397 \\
        CMTrack & ICIP24 & 48.39 & 64.86 & 62.63 &  44.61  & 61.47 & 52.90 & 1467 & 3821 \\
        
        Hybrid-SORT & AAAI24 & 47.87 & 60.77 & 61.07 & 44.18 & \textbf{\textcolor{blue}{63.48}} & 52.25 & 1451 & 2397 \\
         TOPICTrack & TIP25 & 37.71 & 61.33 & 46.54 & 27.39 & 37.22 & 52.24 & 1814 & 4929 \\
         TrackTrack & CVPR25 & \textbf{\textcolor{blue}{51.66}} & \textbf{\textcolor{blue}{72.48}} & \textbf{\textcolor{blue}{67.25}} & 45.87 & 63.17& 56.23 & \textbf{1189} & \textbf{1807} \\
        \hline
         \rowcolor{gray!20} %
        SMTrack & Ours &  \textbf{52.62}	& \textbf{75.07} 	& \textbf{68.15}&	 \textbf{47.87}	& \textbf{64.00} &	 \textbf{61.34}	& \textbf{\textcolor{blue}{1267}}	& \textbf{\textcolor{blue}{1984}}
         \\
        \hline
    \end{tabular}

    \label{tab:BEE24}
\end{table*}

\subsubsection{SpikeYOLOX Head} To enable efficient spiking neural network-based detection for MOT scenarios, we introduce an enhanced detection head inspired by YOLOX, referred to as the SpikeYOLOX Head. This module adopts a decoupled head architecture, separating classification and regression branches, and expands to three detection scales to facilitate multi-scale object modeling and temporal information encoding. As illustrated in Figure~\ref{fig:SpikeYOLOXHead}, the SpikeYOLOX Head is composed of three key components integrated into a unified architecture. First, a feature transformation layer implemented with SpikeConv is used to align the channel dimensions of the backbone outputs. This is followed by two parallel branches: a classification branch, which consists of two SpikeConv layers and a final SpikeConvWithoutBN to produce class probability scores; and a regression branch, which also contains two SpikeConv layers and connects to two independent prediction heads responsible for generating the bounding box coordinates and objectness scores, respectively. This architecture allows task-specific processing in a decoupled manner while preserving temporal dynamics through spiking computations.

\subsubsection{Asa-NWDLoss} While the IoU loss is widely used in object detection, it is highly sensitive to localization shifts for small objects. The NWDLoss \cite{NWDloss} can alleviate the adverse effects of small object scales on the loss function, but existing approaches typically use the global average absolute size across the entire training set as the normalization factor. This strategy is susceptible to being influenced by large object sizes in multi-scale scenarios, leading to an excessively large normalization factor, which in turn weakens the model’s discriminative power and training sensitivity for small objects. To enhance localization robustness under diverse object scales, we propose an an adaptive and scale-aware Normalized Wasserstein Distance loss (Asa-NWDLoss). The method computes the average object size within each training batch and dynamically adjusts the normalization factor, thereby enhancing sensitivity to small objects. Specifically, we model bounding boxes as 2D Gaussian distributions based on their center positions and normalized width/height parameters. The similarity between two such distributions is computed using the second-order Wasserstein distance, which measures both positional and shape dissimilarities. Given two Gaussian distributions \( \mu_1 = \mathbb{N}(m_1, \Sigma_1) \) and \( \mu_2 = \mathbb{N}(m_2, \Sigma_2) \), the second-order Wasserstein distance is expressed as:

\begin{equation}
W_2^2(\mu_1, \mu_2) = \| m_1 - m_2 \|_2^2 + \| \Sigma_1^{1/2} - \Sigma_2^{1/2} \|_F^2,
\label{eq:w2_distance}
\end{equation}

where \( \| \cdot \|_F \) denotes the Frobenius norm. When applying this to bounding boxes, we define the mean and covariance directly from the center coordinates and half-size of the box, enabling a closed-form expression for the distance.

However, since Wasserstein distance is a dissimilarity measure, we further apply an exponential transformation to convert it into a normalized similarity score within the range \([0, 1]\), similar to IoU:

\begin{equation}
\mathrm{NWD}(\mu_1, \mu_2) = \exp\left( -\sqrt{ \frac{W_2^2(\mu_1, \mu_2)}{C_b} } \right),
\label{eq:nwd}
\end{equation}
Here, the normalization factor \( C_{\text{b}} \) is dynamically computed per batch based on the average object size:

\begin{equation}
C_{\text{b}} = \lambda \cdot \sqrt{ \frac{1}{N} \sum_{i=1}^{N} w_i \cdot h_i },
\label{eq:cbatch}
\end{equation}

\( N \) is the number of ground truth boxes in the batch, and \( w_i, h_i \) are the width and height of the \( i \)-th object. The scalar \( \lambda \)=0.8 is a tunable scaling factor.

\subsection{Tracking Pipeline}
Our SMTrack leverages the online multi-object tracking pipeline of TrackTrack \cite{TrackTrack}. TrackTrack is designed around two core strategies: Track-Perspective-Based Association (TPA), which focuses on associating each track with the most suitable detection by fully utilizing high-confidence, low-confidence, and even NMS-suppressed detections, and Track-Aware Initialization (TAI), which suppresses redundant track creation by considering overlaps with active tracks and more confident detections. Together, these mechanisms prioritize long-term, stable tracks and improve robustness against occlusion and detection noise. In SMTrack, SpikeYOLOX supplies the candidate detections, while TrackTrack handles all subsequent motion prediction (via the NSA Kalman Filter) and identity association, the entire association process remains training-free, ensuring efficient and robust identity matching, thereby establishing a solid foundation for SNN-based multi-object tracking.

\begin{table*}[h]
    \centering
    \caption{Comparison with state-of-the-art methods on the MOT17 test set. Metrics marked with $\uparrow$ indicate that higher values are better, and vice versa. Bold and blue values represent the best and second-best results in each column, respectively.}
    \renewcommand{\arraystretch}{1.3} 
    \begin{tabular}{lccccccccc}

        \hline
        \textbf{Tracker} & \textbf{Venue} & \textbf{HOTA $\uparrow$} & \textbf{MOTA $\uparrow$} & \textbf{IDF1 $\uparrow$} & \textbf{AssA $\uparrow$} & \textbf{AssR $\uparrow$}  & \textbf{DetA $\uparrow$} & \textbf{IDs $\downarrow$} & \textbf{Frag $\downarrow$} \\
        \hline
    ByteTrack &	ECCV22 & 58.9 &	75.3 & 70.4 & 56.5 & 61.4 & 61.6 & 2163& 3522\\
    OC-SORT & CVPR23 & 57.1 & 72.7 & 70.0 & 56.0 & 61.1 & 58.5 & 2997 & 5652\\
    Deep-OC-SORT &ICIP23& 59.4 & 76.3 & 71.9 & 57.5 & 62.5 & 61.5& 2388 &3186\\
 AdapTrack & ICIP24 & 60.7 & 74.2 & 74.7 & 60.8 & 67.1 & 60.8 & 2022 & 2268\\

    CMTrack & ICIP24 & 60.0 & 75.3 & 74.1 & 59.4 & 66.3 & 60.9 & 1416 & 1914 \\
    Hybrid-SORT & AAAI24 & 59.5& 74.6& 73.6 & 58.8 & 64.6 & 60.5 & 2733 & 5130\\
    TOPICTrack & TIP25 & 60.0	& 76.8 & 71.8 & 58.2 & 62.6 & 62.1 & 1824 & 2823\\
    TrackTrack & CVPR25 & \textbf{\textcolor{blue}{62.0}} & \textbf{77.4} & \textbf{\textcolor{blue}{76.0}} & \textbf{\textcolor{blue}{62.7}} & \textbf{\textcolor{blue}{67.8}} & \textbf{\textcolor{blue}{62.7}} & \textbf{\textcolor{blue}{1395}} & \textbf{\textcolor{blue}{1959}}\\
        \hline
         \rowcolor{gray!20} %
        SMTrack & Ours & \textbf{63.9} & \textbf{\textcolor{blue}{76.9}} & \textbf{78.3} & \textbf{64.5} & \textbf{70.3} & \textbf{63.6} & \textbf{1032} & \textbf{1704} \\
        \hline
    \end{tabular}

    \label{tab:MOT17}
\end{table*}

\begin{table*}[h]
    \centering
    \caption{Comparison with state-of-the-art methods on the MOT20 test set. Metrics marked with $\uparrow$ indicate that higher values are better, and vice versa. Bold and blue values represent the best and second-best results in each column, respectively.}
    \renewcommand{\arraystretch}{1.3} 
    \begin{tabular}{lccccccccc}

        \hline
        \textbf{Tracker} & \textbf{Venue} & \textbf{HOTA $\uparrow$} & \textbf{MOTA $\uparrow$} & \textbf{IDF1 $\uparrow$} & \textbf{AssA $\uparrow$} & \textbf{AssR $\uparrow$}  & \textbf{DetA $\uparrow$} & \textbf{IDs $\downarrow$} & \textbf{Frag $\downarrow$} \\
        \hline
        ByteTrack & ECCV22 & 52.6 & 68.1 & 63.6 & 48.8 & 56.0 & 57.0 & 2946 & 4202 \\

       OC-SORT	&CVPR23 & 53.2 & 64.5 & 66.1 & 53.4 & 57.7 & 53.1 &1532 & 5168\\

        Deep OC-SORT & ICIP23 & 54.7 &  \textbf{\textcolor{blue}{70.0}}  & 66.4 & 51.2 & 55.8  & \textbf{58.5} & 2116 & 2728 \\
        AdapTrack & ICIP24 & 57.5 &	59.7 & 71.1 & 61.3 & 67.0 &	54.2 & 2704 & 1903\\

        CMTrack & ICIP24 & 57.4 & 66.2 & 69.8 & 58.2 & 62.2 & 56.9 & 1861 & 1533 \\
        Hybrid-SORT  &	AAAI24 &	55.7 &	64.7 &	67.3	 &54.9 &	59.9	 &	56.8	 &3547	 &5996\\
         TOPICTrack & TIP25 & 54.0 & 69.6 & 65.9 & 51.5 & 55.9  & 56.7 & 1461 & 2119 \\
        TrackTrack & CVPR25 &  \textbf{\textcolor{blue}{59.5}} & 69.5 & \textbf{\textcolor{blue}{73.7}} & \textbf{\textcolor{blue}{61.4}} & \textbf{\textcolor{blue}{65.4}}  &  \textbf{\textcolor{blue}{58.2}}  & \textbf{\textcolor{blue}{1307}} & \textbf{\textcolor{blue}{1432}} \\
        \hline
         \rowcolor{gray!20} %
        SMTrack & Ours & \textbf{59.9} & \textbf{70.5} & \textbf{74.8} & \textbf{61.5} & \textbf{65.5} &  \textbf{58.5} & \textbf{1120} & \textbf{1430} \\
        \hline
    \end{tabular}
    
    \label{tab:MOT20}
\end{table*}
\section{Evaluation}
\subsection{Experimental Setting}
\subsubsection {Datasets} To comprehensively evaluate multi-object tracking performance, this study employs four public benchmark datasets with distinct challenges: MOT17 \cite{MOT17}, MOT20 \cite{MOT20}, DanceTrack \cite{dancetrack}, and BEE24 \cite{TOPICTrack}. MOT17 and MOT20 focus on pedestrian tracking. MOT17 covers diverse surveillance scenarios with both static and dynamic camera views of crowd activities, while MOT20 concentrates on extreme crowding, posing severe occlusion challenges. DanceTrack involves dance scenes with complex motion patterns and significant variations. BEE24 presents dense bee swarm scenarios, focusing on tracking small, visually similar objects with complex movements, including instant transitions from stationary to flying. The diversity of these datasets ensures a thorough evaluation of tracking algorithms across various real-world applications.

\subsubsection {Metrics} We evaluate tracker performance using several standard metrics, including HOTA  \cite{Hota}, CLEAR \cite{MOTA}, and IDF1 \cite{IDF1}. HOTA, considered the most balanced and comprehensive metric, jointly assesses detection, association, and localization accuracy, and has been shown to align well with human visual perception. MOTA, part of the CLEAR metric suite, captures overall tracking errors by integrating false positives, false negatives, and identity switches. IDF1 evaluates the consistency of identity assignment across frames, emphasizing accurate identity preservation. Following common practice in prior studies \cite{smiletrack,ucmctrack,HybridSort}, we adopt HOTA as the primary metric for assessing overall multi-object tracking performance.

\subsubsection{Trackers} We conduct a comprehensive evaluation of our SMTrack against eight state-of-the-art tracking methods, including OC-SORT \cite{OCSort}, AdapTrack \cite{Adaptrack}, CMTrack \cite{CMTrack}, ByteTrack \cite{ByteTrack}, TOPICTrack \cite{TOPICTrack}, Deep-OC-SORT \cite{DeepOCsort}, TrackTrack \cite{TrackTrack}, and Hybrid-SORT \cite{HybridSort}.

\subsubsection{Implementation Details}
For object detection, all ANN-based methods adopted the same YOLOX-s model \cite{yolox}, while our SNN-based model, SMTrack, shared a similar configuration. We initialized from COCO \cite{COCO} pretrained weights and trained on each target dataset individually. Specifically, MOT17 \cite{MOT17} and MOT20 \cite{MOT20} were trained for 80 epochs following the default training settings from \cite{ByteTrack}, while DanceTrack and BEE24 were trained for 120 epochs according to the settings in \cite{TOPICTrack}. SMTrack followed the same training strategy as the ANN-based methods, using a YOLOX-s scale (depth = 0.33, width = 0.50). In SMTrack, the hyperparameter \( D \), which defines the maximum integer value emitted by the I-LIF neuron, is set to 4. The feature extractor was configured identically to TrackTrack \cite{TrackTrack}. All experiments were conducted on a platform equipped with an NVIDIA Titan X GPU, 16GB RAM, and an Intel i9-10850K processor (3.6GHz).

\subsection{Benchmark Results}
In this section, we present the quantitative evaluation results of the proposed SMTrack on four benchmark datasets: DanceTrack \cite{dancetrack}, BEE24 \cite{TOPICTrack}, MOT17 \cite{MOT17}, and MOT20 \cite{MOT20}. As shown in Tables~\ref{tab:DanceTrack}–\ref{tab:MOT20} , SMTrack consistently outperforms all state-of-the-art trackers across all datasets. These results demonstrate that SMTrack not only achieves comparable overall performance to leading ANN-based MOT methods, but also surpasses them in several key metrics, offering a robust and accurate SNN-based solution for multi-object tracking in complex and dynamic scenarios.

\subsubsection{DanceTrack} Table \ref{tab:DanceTrack} presents the evaluation results of SMTrack and eight state-of-the-art trackers on the DanceTrack dataset. SMTrack demonstrates leading performance across several key metrics. Specifically, it achieves a HOTA score of 57.76, surpassing the baseline tracker TrackTrack (56.54) by 1.22 points. Moreover, SMTrack achieves the highest scores in IDF1 (58.47), AssA (42.62), and DetA (47.91), highlighting its strong capability in both identity association accuracy and detection alignment. Although SMTrack does not achieve the highest MOTA score, its result of 89.62 is only 0.39 points lower than Deep OC-SORT, which holds the top score of 90.01. Importantly, SMTrack achieves the lowest number of identity switches (IDs = 857) among all methods, suggesting excellent identity consistency even in the presence of frequent occlusions and dynamic interactions.

\subsubsection{BEE24} Table \ref{tab:BEE24} presents the comparison results on the BEE24 test set. SMTrack demonstrates leading performance across all major evaluation metrics, indicating its strong capability in handling complex small-object tracking scenarios with severe occlusions and dense object interactions. Specifically, SMTrack achieves a HOTA score of 52.62, significantly outperforming the second-best method, TrackTrack (51.66), and consistently surpasses all competing methods in other key metrics such as MOTA (75.07) and IDF1 (68.15). Although SMTrack does not achieve the lowest values in identity switches (IDs = 1267) and fragmentation (Frag = 1984), it ranks second in both metrics, closely following TrackTrack. This indicates that SMTrack maintains excellent temporal consistency and identity stability, even under challenging dynamic conditions.
\begin{table}[h]
    \centering
    \caption{Performance of SMTrack under different spike time-step configurations on MOT17 and BEE24 datasets.}
    \renewcommand{\arraystretch}{1.3}
    \resizebox{0.48\textwidth}{!}{
    \begin{tabular}{lccccc}
        \hline
        \textbf{DateSet} & \textbf{T} & \textbf{HOTA $\uparrow$} & \textbf{IDF1 $\uparrow$} & \textbf{MOTA $\uparrow$} & \textbf{DetA $\uparrow$} \\
        \hline
       \multirow{4}{*}{\textbf{BEE24}} 
      
        & 1 & 48.61 & 63.59 & 67.66 & 53.21 \\
        & 2 & 52.53 & 67.99 & 75.08 & 61.32 \\
        
        & 4 & 52.62 & 68.15 & 75.07 & 61.34 \\
        & 8 & \textbf{52.93} & \textbf{68.99} & \textbf{75.18} & \textbf{61.46} \\
        \hline
        \multirow{4}{*}{\textbf{MOT17}}  
        &  1 & 79.84 & 90.53 & 90.86 & 80.45 \\
        &  2 & 81.91 & 91.70 & 92.21 & 83.38 \\
        &  4 & 82.09 & 91.66 &  93.48 & 83.45\\
         & 8 & \textbf{82.64} & \textbf{92.56} & \textbf{92.79} & \textbf{82.93} \\
        \hline
    \end{tabular}
    }

    \label{tab:ablation-timing}
\end{table}

\begin{table}[h] 
\centering
\renewcommand{\arraystretch}{1.3}
\caption{Comparison of different $\lambda$ values in Asa-NWDLoss on MOT17 (D=4, T=2). The bottom row uses a fixed dataset-wide average size as normalization.}
\resizebox{0.48\textwidth}{!}{
\begin{tabular}{c|cccc}
\hline
$\lambda$ & HOTA $\uparrow$ & MOTA $\uparrow$ & IDF1 $\uparrow$ & DetA $\uparrow$ \\
\hline
0.2 & 80.12 & 89.81 & 91.90 & 78.88 \\
0.4 & 80.94 & 91.98 & 91.15 & 81.57 \\
0.6 & 81.01 & 90.49 & 90.53 & 80.58 \\
0.8 & \textbf{81.91} & \textbf{91.70} & \textbf{91.21} & \textbf{83.38} \\
1.0 & 81.29 & 91.24 & 91.21 & 82.11 \\

\hline
Fixed $C$ & 78.29 & 90.81 & 78.19 & 78.59 \\
\hline
\end{tabular}
}
\label{tab:lambda_mot17}
\end{table}

\subsubsection{MOT20} Table \ref{tab:MOT20} shows that SMTrack achieves outstanding performance across all core evaluation metrics. Specifically, it attains the highest HOTA score of 59.9 among all evaluated methods, along with a MOTA of 70.5 and an IDF1 of 74.8, demonstrating excellent detection accuracy and strong identity preservation. Compared to the best-performing ANN-based tracker, TrackTrack, SMTrack achieves improvements in all three metrics, indicating that the SNN-based architecture provides high-quality and temporally consistent detections for MOT tasks. In terms of association quality, SMTrack also excels, achieving leading scores in AssA (61.5) and AssR (65.5), which reflect its high reliability in cross-frame object association. More importantly, it records the lowest number of identity switches (1120) and trajectory fragments (1430), further validating its robustness and temporal consistency under challenging occlusion-heavy scenarios.

\subsection{Ablation Studies}

\subsubsection{Effect of Spike Time Steps}

To evaluate the impact of spike time-step configurations on tracking performance, we conduct ablation experiments on both the BEE24 and MOT17 datasets with $T \in \{1, 2, 4, 8\}$. The results are summarized in Table~\ref{tab:ablation-timing}. We observe that increasing the number of time steps generally improves performance across all metrics. On BEE24, HOTA improves from 48.61 at $T=1$ to 52.93 at $T=8$, while on MOT17, HOTA increases from 79.84 to 82.64. This suggests that longer spike durations allow neurons to accumulate more temporal information, leading to more stable activations, better detection accuracy, and improved temporal consistency. However, longer time steps also incur higher computational latency and power consumption, which contradicts the low-latency, energy-efficient design goals of spiking systems. To balance accuracy and efficiency, we select $T=2$ as the default configuration for SMTrack. This setting achieves near-optimal performance on both datasets while maintaining low latency, making it better suited for deployment in resource-constrained environments.


\subsubsection{Effect of $\lambda$ in Asa-NWDLoss}
To evaluate the impact of the scaling coefficient $\lambda$ in Asa-NWDLoss, we conduct experiments on the MOT17 dataset under a fixed configuration (D=4, T=2), as shown in Table~\ref{tab:lambda_mot17}. The parameter $\lambda$ controls the strength of the normalization factor, which is computed from the average object size within each batch. We observe that performance generally improves as $\lambda$ increases from 0.2 to 0.8, peaking at $\lambda=0.8$ with the highest HOTA (81.91), MOTA (91.70), IDF1 (91.21), and DetA (83.38). This indicates that appropriately scaling the batch-specific normalization factor allows the loss to better balance small and large object sensitivity. When $\lambda$ is too small, under-normalization reduces sensitivity to small-scale variation; when it is too large, over-normalization may suppress useful gradients. Additionally, the last row uses a fixed constant $C$ derived from the dataset-wide average size, as done in prior work~\cite{NWDloss}. Compared to the adaptive strategy, this baseline results in noticeably lower HOTA (78.29) and IDF1 (78.19), confirming the benefit of batch-aware dynamic adjustment in our Asa-NWDLoss formulation.

\section{Conclusion}
In this paper, we present SMTrack, the first directly-trained deep spiking neural network (SNN) framework for end-to-end multi-object tracking (MOT) on conventional RGB videos. To further enhance regression robustness under multi-scale scenarios, we propose a novel Asa-NWDLoss, an adaptive and scale-aware Normalized Wasserstein Distance loss function that dynamically adjusts the normalization factor based on batch-level object sizes. We also integrate the state-aware TrackTrack identity association module to ensure robust and coherent tracking under crowded or occluded conditions. Extensive experiments on BEE24, MOT17, MOT20, and DanceTrack demonstrate that SMTrack achieves state-of-the-art SNN-based tracking performance, and even competes with leading ANN-based trackers—all while retaining the energy efficiency of spike-based computation. These results highlight the promise of SNNs for real-world multi-object tracking applications beyond neuromorphic inputs.

\bibliographystyle{IEEEtran}
\bibliography{main}

\begin{thebibliography}{10}
\providecommand{\url}[1]{#1}
\csname url@samestyle\endcsname
\providecommand{\newblock}{\relax}
\providecommand{\bibinfo}[2]{#2}
\providecommand{\BIBentrySTDinterwordspacing}{\spaceskip=0pt\relax}
\providecommand{\BIBentryALTinterwordstretchfactor}{4}
\providecommand{\BIBentryALTinterwordspacing}{\spaceskip=\fontdimen2\font plus
\BIBentryALTinterwordstretchfactor\fontdimen3\font minus \fontdimen4\font\relax}
\providecommand{\BIBforeignlanguage}[2]{{%
\expandafter\ifx\csname l@#1\endcsname\relax
\typeout{** WARNING: IEEEtran.bst: No hyphenation pattern has been}%
\typeout{** loaded for the language `#1'. Using the pattern for}%
\typeout{** the default language instead.}%
\else
\language=\csname l@#1\endcsname
\fi
#2}}
\providecommand{\BIBdecl}{\relax}
\BIBdecl

\bibitem{oh2011large}
S.~Oh, A.~Hoogs, A.~Perera, N.~Cuntoor, C.-C. Chen, J.~T. Lee, S.~Mukherjee, J.~K. Aggarwal, H.~Lee, L.~Davis \emph{et~al.}, ``A large-scale benchmark dataset for event recognition in surveillance video,'' in \emph{CVPR 2011}.\hskip 1em plus 0.5em minus 0.4em\relax IEEE, 2011, pp. 3153--3160.

\bibitem{sportsmot}
Y.~Cui, C.~Zeng, X.~Zhao, Y.~Yang, G.~Wu, and L.~Wang, ``Sportsmot: A large multi-object tracking dataset in multiple sports scenes,'' in \emph{Proceedings of the IEEE/CVF international conference on computer vision}, 2023, pp. 9921--9931.

\bibitem{Deepdriving}
C.~Chen, A.~Seff, A.~Kornhauser, and J.~Xiao, ``Deepdriving: Learning affordance for direct perception in autonomous driving,'' in \emph{Proceedings of the IEEE international conference on computer vision}, 2015, pp. 2722--2730.

\bibitem{zhang2019robust}
W.~Zhang, H.~Zhou, S.~Sun, Z.~Wang, J.~Shi, and C.~C. Loy, ``Robust multi-modality multi-object tracking,'' in \emph{Proceedings of the IEEE/CVF international conference on computer vision}, 2019, pp. 2365--2374.

\bibitem{xiang2015learning}
Y.~Xiang, A.~Alahi, and S.~Savarese, ``Learning to track: Online multi-object tracking by decision making,'' in \emph{Proceedings of the IEEE international conference on computer vision}, 2015, pp. 4705--4713.

\bibitem{ByteTrack}
Y.~Zhang, P.~Sun, Y.~Jiang, D.~Yu, F.~Weng, Z.~Yuan, P.~Luo, W.~Liu, and X.~Wang, ``Bytetrack: Multi-object tracking by associating every detection box,'' in \emph{European conference on computer vision}.\hskip 1em plus 0.5em minus 0.4em\relax Springer, 2022, pp. 1--21.

\bibitem{tu2021resource}
J.~Tu, C.~Chen, Q.~Xu, B.~Yang, and X.~Guan, ``Resource-efficient visual multiobject tracking on embedded device,'' \emph{IEEE Internet of Things Journal}, vol.~9, no.~11, pp. 8531--8543, 2021.

\bibitem{li2023multiobject}
H.~Li, X.~Liang, H.~Yin, L.~Xu, X.~Kong, and T.~A. Gulliver, ``Multiobject tracking via discriminative embeddings for the internet of things,'' \emph{IEEE Internet of Things Journal}, vol.~10, no.~12, pp. 10\,532--10\,546, 2023.

\bibitem{maass1997networks}
W.~Maass, ``Networks of spiking neurons: the third generation of neural network models,'' \emph{Neural networks}, vol.~10, no.~9, pp. 1659--1671, 1997.

\bibitem{roy2019towards}
K.~Roy, A.~Jaiswal, and P.~Panda, ``Towards spike-based machine intelligence with neuromorphic computing,'' \emph{Nature}, vol. 575, no. 7784, pp. 607--617, 2019.

\bibitem{merolla2014million}
P.~A. Merolla, J.~V. Arthur, R.~Alvarez-Icaza, A.~S. Cassidy, J.~Sawada, F.~Akopyan, B.~L. Jackson, N.~Imam, C.~Guo, Y.~Nakamura \emph{et~al.}, ``A million spiking-neuron integrated circuit with a scalable communication network and interface,'' \emph{Science}, vol. 345, no. 6197, pp. 668--673, 2014.

\bibitem{poon2011neuromorphic}
C.-S. Poon and K.~Zhou, ``Neuromorphic silicon neurons and large-scale neural networks: challenges and opportunities,'' \emph{Frontiers in neuroscience}, vol.~5, p. 108, 2011.

\bibitem{deng2022temporal}
S.~Deng, Y.~Li, S.~Zhang, and S.~Gu, ``Temporal efficient training of spiking neural network via gradient re-weighting,'' \emph{arXiv preprint arXiv:2202.11946}, 2022.

\bibitem{EM-ResNet}
W.~Fang, Z.~Yu, Y.~Chen, T.~Huang, T.~Masquelier, and Y.~Tian, ``Deep residual learning in spiking neural networks,'' \emph{Advances in Neural Information Processing Systems}, vol.~34, pp. 21\,056--21\,069, 2021.

\bibitem{guo2023rmp}
Y.~Guo, X.~Liu, Y.~Chen, L.~Zhang, W.~Peng, Y.~Zhang, X.~Huang, and Z.~Ma, ``Rmp-loss: Regularizing membrane potential distribution for spiking neural networks,'' in \emph{Proceedings of the IEEE/CVF International Conference on Computer Vision}, 2023, pp. 17\,391--17\,401.

\bibitem{Spiking-yolo}
S.~Kim, S.~Park, B.~Na, and S.~Yoon, ``Spiking-yolo: spiking neural network for energy-efficient object detection,'' in \emph{Proceedings of the AAAI conference on artificial intelligence}, vol.~34, no.~07, 2020, pp. 11\,270--11\,277.

\bibitem{Meta-SpikeFormer}
M.~Yao, J.~Hu, T.~Hu, Y.~Xu, Z.~Zhou, Y.~Tian, B.~Xu, and G.~Li, ``Spike-driven transformer v2: Meta spiking neural network architecture inspiring the design of next-generation neuromorphic chips,'' \emph{arXiv preprint arXiv:2404.03663}, 2024.

\bibitem{EMS-YOLO}
Q.~Su, Y.~Chou, Y.~Hu, J.~Li, S.~Mei, Z.~Zhang, and G.~Li, ``Deep directly-trained spiking neural networks for object detection. 2023 ieee,'' in \emph{CVF International Conference on Computer Vision (ICCV)}, 2023, pp. 6532--6542.

\bibitem{SpikeYOLO}
X.~Luo, M.~Yao, Y.~Chou, B.~Xu, and G.~Li, ``Integer-valued training and spike-driven inference spiking neural network for high-performance and energy-efficient object detection,'' in \emph{European Conference on Computer Vision}.\hskip 1em plus 0.5em minus 0.4em\relax Springer, 2024, pp. 253--272.

\bibitem{mitrokhin2018event}
A.~Mitrokhin, C.~Ferm{\"u}ller, C.~Parameshwara, and Y.~Aloimonos, ``Event-based moving object detection and tracking,'' in \emph{2018 IEEE/RSJ International Conference on Intelligent Robots and Systems (IROS)}.\hskip 1em plus 0.5em minus 0.4em\relax IEEE, 2018, pp. 1--9.

\bibitem{STNet}
J.~Zhang, B.~Dong, H.~Zhang, J.~Ding, F.~Heide, B.~Yin, and X.~Yang, ``Spiking transformers for event-based single object tracking,'' in \emph{Proceedings of the IEEE/CVF conference on Computer Vision and Pattern Recognition}, 2022, pp. 8801--8810.

\bibitem{zheng2022spike}
Y.~Zheng, Z.~Yu, S.~Wang, and T.~Huang, ``Spike-based motion estimation for object tracking through bio-inspired unsupervised learning,'' \emph{IEEE Transactions on Image Processing}, vol.~32, pp. 335--349, 2022.

\bibitem{qu2024spike}
J.~Qu, Z.~Gao, Y.~Li, Y.~Lu, and H.~Qiao, ``Spike-based high energy efficiency and accuracy tracker for robot,'' in \emph{2024 IEEE/RSJ International Conference on Intelligent Robots and Systems (IROS)}.\hskip 1em plus 0.5em minus 0.4em\relax IEEE, 2024, pp. 1428--1434.

\bibitem{NeuroSORT}
Z.~Shen, X.~Xie, C.~Fang, F.~Tian, S.~Ma, J.~Yang, and M.~Sawan, ``Neurosort: A neuromorphic accelerator for spike-based online and real-time tracking,'' in \emph{2024 IEEE 6th International Conference on AI Circuits and Systems (AICAS)}.\hskip 1em plus 0.5em minus 0.4em\relax IEEE, 2024, pp. 312--316.

\bibitem{SpikeCalib}
Y.~Li, X.~He, Y.~Dong, Q.~Kong, and Y.~Zeng, ``Spike calibration: Fast and accurate conversion of spiking neural network for object detection and segmentation. arxiv 2022,'' \emph{arXiv preprint arXiv:2207.02702}, 2022.

\bibitem{MSResNet}
Y.~Hu, L.~Deng, Y.~Wu, M.~Yao, and G.~Li, ``Advancing spiking neural networks toward deep residual learning,'' \emph{IEEE transactions on neural networks and learning systems}, vol.~36, no.~2, pp. 2353--2367, 2024.

\bibitem{COCO}
T.-Y. Lin, M.~Maire, S.~Belongie, J.~Hays, P.~Perona, D.~Ramanan, P.~Doll{\'a}r, and C.~L. Zitnick, ``Microsoft coco: Common objects in context,'' in \emph{Computer vision--ECCV 2014: 13th European conference, zurich, Switzerland, September 6-12, 2014, proceedings, part v 13}.\hskip 1em plus 0.5em minus 0.4em\relax Springer, 2014, pp. 740--755.

\bibitem{Gen1}
P.~De~Tournemire, D.~Nitti, E.~Perot, D.~Migliore, and A.~Sironi, ``A large scale event-based detection dataset for automotive,'' \emph{arXiv preprint arXiv:2001.08499}, 2020.

\bibitem{iou}
J.~Redmon and A.~Farhadi, ``Yolo9000: better, faster, stronger,'' in \emph{Proceedings of the IEEE conference on computer vision and pattern recognition}, 2017, pp. 7263--7271.

\bibitem{NWDloss}
J.~Wang, C.~Xu, W.~Yang, and L.~Yu, ``A normalized gaussian wasserstein distance for tiny object detection,'' \emph{arXiv preprint arXiv:2110.13389}, 2021.

\bibitem{yolox}
Z.~Ge, S.~Liu, F.~Wang, Z.~Li, and J.~Sun, ``Yolox: Exceeding yolo series in 2021,'' \emph{arXiv preprint arXiv:2107.08430}, 2021.

\bibitem{TrackTrack}
K.~Shim, K.~Ko, Y.~Yang, and C.~Kim, ``Focusing on tracks for online multi-object tracking,'' in \emph{Proceedings of the Computer Vision and Pattern Recognition Conference}, 2025, pp. 11\,687--11\,696.

\bibitem{liu2022ultralow}
Q.~Liu and Z.~Zhang, ``Ultralow power always-on intelligent and connected snn-based system for multimedia iot-enabled applications,'' \emph{IEEE Internet of Things Journal}, vol.~9, no.~17, pp. 15\,570--15\,577, 2022.

\bibitem{Seenn}
Y.~Li, T.~Geller, Y.~Kim, and P.~Panda, ``Seenn: towards temporal spiking early exit neural networks,'' \emph{Advances in Neural Information Processing Systems}, vol.~36, pp. 63\,327--63\,342, 2023.

\bibitem{YOLO}
J.~Redmon, S.~Divvala, R.~Girshick, and A.~Farhadi, ``You only look once: Unified, real-time object detection,'' in \emph{Proceedings of the IEEE conference on computer vision and pattern recognition}, 2016, pp. 779--788.

\bibitem{bewley2016simple}
A.~Bewley, Z.~Ge, L.~Ott, F.~Ramos, and B.~Upcroft, ``Simple online and realtime tracking,'' in \emph{2016 IEEE international conference on image processing (ICIP)}.\hskip 1em plus 0.5em minus 0.4em\relax Ieee, 2016, pp. 3464--3468.

\bibitem{StrongSORT}
Y.~Du, Z.~Zhao, Y.~Song, Y.~Zhao, F.~Su, T.~Gong, and H.~Meng, ``Strongsort: Make deepsort great again,'' \emph{IEEE Transactions on Multimedia}, vol.~25, pp. 8725--8737, 2023.

\bibitem{he2021learnable}
J.~He, Z.~Huang, N.~Wang, and Z.~Zhang, ``Learnable graph matching: Incorporating graph partitioning with deep feature learning for multiple object tracking,'' in \emph{Proceedings of the IEEE/CVF conference on computer vision and pattern recognition}, 2021, pp. 5299--5309.

\bibitem{QDTrack}
J.~Pang, L.~Qiu, X.~Li, H.~Chen, Q.~Li, T.~Darrell, and F.~Yu, ``Quasi-dense similarity learning for multiple object tracking,'' in \emph{Proceedings of the IEEE/CVF conference on computer vision and pattern recognition}, 2021, pp. 164--173.

\bibitem{Hungarian}
H.~W. Kuhn, ``The hungarian method for the assignment problem,'' \emph{Naval Research Logistics (NRL)}, vol.~52, no.~1, pp. 7--21, 2004.

\bibitem{Mat}
S.~Han, P.~Huang, H.~Wang, E.~Yu, D.~Liu, and X.~Pan, ``Mat: Motion-aware multi-object tracking,'' \emph{Neurocomputing}, vol. 476, pp. 75--86, 2022.

\bibitem{khurana2021detecting}
T.~Khurana, A.~Dave, and D.~Ramanan, ``Detecting invisible people,'' in \emph{Proceedings of the IEEE/CVF international conference on computer vision}, 2021, pp. 3174--3184.

\bibitem{Motiontrack}
Z.~Qin, S.~Zhou, L.~Wang, J.~Duan, G.~Hua, and W.~Tang, ``Motiontrack: Learning robust short-term and long-term motions for multi-object tracking,'' in \emph{Proceedings of the IEEE/CVF conference on computer vision and pattern recognition}, 2023, pp. 17\,939--17\,948.

\bibitem{wojke2017simple}
N.~Wojke, A.~Bewley, and D.~Paulus, ``Simple online and realtime tracking with a deep association metric,'' in \emph{2017 IEEE international conference on image processing (ICIP)}.\hskip 1em plus 0.5em minus 0.4em\relax IEEE, 2017, pp. 3645--3649.

\bibitem{kim2021discriminative}
C.~Kim, L.~Fuxin, M.~Alotaibi, and J.~M. Rehg, ``Discriminative appearance modeling with multi-track pooling for real-time multi-object tracking,'' in \emph{Proceedings of the IEEE/CVF conference on computer vision and pattern recognition}, 2021, pp. 9553--9562.

\bibitem{geiger2013vision}
A.~Geiger, P.~Lenz, C.~Stiller, and R.~Urtasun, ``Vision meets robotics: The kitti dataset,'' \emph{The international journal of robotics research}, vol.~32, no.~11, pp. 1231--1237, 2013.

\bibitem{Bot-sort}
N.~Aharon, R.~Orfaig, and B.-Z. Bobrovsky, ``Bot-sort: Robust associations multi-pedestrian tracking,'' \emph{arXiv preprint arXiv:2206.14651}, 2022.

\bibitem{Trackflow}
G.~Mancusi, A.~Panariello, A.~Porrello, M.~Fabbri, S.~Calderara, and R.~Cucchiara, ``Trackflow: Multi-object tracking with normalizing flows,'' in \emph{Proceedings of the IEEE/CVF International Conference on Computer Vision}, 2023, pp. 9531--9543.

\bibitem{JDE}
Z.~Wang, L.~Zheng, Y.~Liu, Y.~Li, and S.~Wang, ``Towards real-time multi-object tracking,'' in \emph{European conference on computer vision}.\hskip 1em plus 0.5em minus 0.4em\relax Springer, 2020, pp. 107--122.

\bibitem{Mots}
P.~Voigtlaender, M.~Krause, A.~Osep, J.~Luiten, B.~B.~G. Sekar, A.~Geiger, and B.~Leibe, ``Mots: Multi-object tracking and segmentation,'' in \emph{Proceedings of the ieee/cvf conference on computer vision and pattern recognition}, 2019, pp. 7942--7951.

\bibitem{Fairmot}
Y.~Zhang, C.~Wang, X.~Wang, W.~Zeng, and W.~Liu, ``Fairmot: On the fairness of detection and re-identification in multiple object tracking,'' \emph{International journal of computer vision}, vol. 129, pp. 3069--3087, 2021.

\bibitem{Transtrack}
P.~Sun, J.~Cao, Y.~Jiang, R.~Zhang, E.~Xie, Z.~Yuan, C.~Wang, and P.~Luo, ``Transtrack: Multiple object tracking with transformer,'' \emph{arXiv preprint arXiv:2012.15460}, 2020.

\bibitem{Trackformer}
T.~Meinhardt, A.~Kirillov, L.~Leal-Taixe, and C.~Feichtenhofer, ``Trackformer: Multi-object tracking with transformers,'' in \emph{Proceedings of the IEEE/CVF conference on computer vision and pattern recognition}, 2022, pp. 8844--8854.

\bibitem{li2023inference}
R.~Li, B.~Zhang, J.~Liu, W.~Liu, and Z.~Teng, ``Inference-domain network evolution: A new perspective for one-shot multi-object tracking,'' \emph{IEEE Transactions on Image Processing}, vol.~32, pp. 2147--2159, 2023.

\bibitem{Qdtrack1}
T.~Fischer, T.~E. Huang, J.~Pang, L.~Qiu, H.~Chen, T.~Darrell, and F.~Yu, ``Qdtrack: Quasi-dense similarity learning for appearance-only multiple object tracking,'' \emph{IEEE Transactions on Pattern Analysis and Machine Intelligence}, vol.~45, no.~12, pp. 15\,380--15\,393, 2023.

\bibitem{dancetrack}
P.~Sun, J.~Cao, Y.~Jiang, Z.~Yuan, S.~Bai, K.~Kitani, and P.~Luo, ``Dancetrack: Multi-object tracking in uniform appearance and diverse motion,'' in \emph{Proceedings of the IEEE/CVF conference on computer vision and pattern recognition}, 2022, pp. 20\,993--21\,002.

\bibitem{Flownet}
A.~Dosovitskiy, P.~Fischer, E.~Ilg, P.~Hausser, C.~Hazirbas, V.~Golkov, P.~Van Der~Smagt, D.~Cremers, and T.~Brox, ``Flownet: Learning optical flow with convolutional networks,'' in \emph{Proceedings of the IEEE international conference on computer vision}, 2015, pp. 2758--2766.

\bibitem{Raft}
Z.~Teed and J.~Deng, ``Raft: Recurrent all-pairs field transforms for optical flow,'' in \emph{Computer Vision--ECCV 2020: 16th European Conference, Glasgow, UK, August 23--28, 2020, Proceedings, Part II 16}.\hskip 1em plus 0.5em minus 0.4em\relax Springer, 2020, pp. 402--419.

\bibitem{matchflow}
Q.~Dong, C.~Cao, and Y.~Fu, ``Rethinking optical flow from geometric matching consistent perspective,'' in \emph{Proceedings of the IEEE/CVF Conference on computer vision and pattern recognition}, 2023, pp. 1337--1347.

\bibitem{TOPICTrack}
X.~Cao, Y.~Zheng, Y.~Yao, H.~Qin, X.~Cao, and S.~Guo, ``Topic: A parallel association paradigm for multi-object tracking under complex motions and diverse scenes,'' \emph{IEEE Transactions on Image Processing}, 2025.

\bibitem{Xception}
F.~Chollet, ``Xception: Deep learning with depthwise separable convolutions,'' in \emph{Proceedings of the IEEE conference on computer vision and pattern recognition}, 2017, pp. 1251--1258.

\bibitem{MOT17}
A.~Milan, L.~Leal-Taix{\'e}, I.~Reid, S.~Roth, and K.~Schindler, ``Mot16: A benchmark for multi-object tracking,'' \emph{arXiv preprint arXiv:1603.00831}, 2016.

\bibitem{MOT20}
P.~Dendorfer, H.~Rezatofighi, A.~Milan, J.~Shi, D.~Cremers, I.~Reid, S.~Roth, K.~Schindler, and L.~Leal-Taix{\'e}, ``Mot20: A benchmark for multi object tracking in crowded scenes,'' \emph{arXiv preprint arXiv:2003.09003}, 2020.

\bibitem{Hota}
J.~Luiten, A.~Osep, P.~Dendorfer, P.~Torr, A.~Geiger, L.~Leal-Taix{\'e}, and B.~Leibe, ``Hota: A higher order metric for evaluating multi-object tracking,'' \emph{International journal of computer vision}, vol. 129, pp. 548--578, 2021.

\bibitem{MOTA}
K.~Bernardin and R.~Stiefelhagen, ``Evaluating multiple object tracking performance: the clear mot metrics,'' \emph{EURASIP Journal on Image and Video Processing}, vol. 2008, pp. 1--10, 2008.

\bibitem{IDF1}
E.~Ristani, F.~Solera, R.~Zou, R.~Cucchiara, and C.~Tomasi, ``Performance measures and a data set for multi-target, multi-camera tracking,'' in \emph{European conference on computer vision}.\hskip 1em plus 0.5em minus 0.4em\relax Springer, 2016, pp. 17--35.

\bibitem{smiletrack}
Y.-H. Wang, J.-W. Hsieh, P.-Y. Chen, M.-C. Chang, H.-H. So, and X.~Li, ``Smiletrack: Similarity learning for occlusion-aware multiple object tracking,'' in \emph{Proceedings of the AAAI Conference on Artificial Intelligence}, vol.~38, no.~6, 2024, pp. 5740--5748.

\bibitem{ucmctrack}
K.~Yi, K.~Luo, X.~Luo, J.~Huang, H.~Wu, R.~Hu, and W.~Hao, ``Ucmctrack: Multi-object tracking with uniform camera motion compensation,'' in \emph{Proceedings of the AAAI conference on artificial intelligence}, vol.~38, no.~7, 2024, pp. 6702--6710.

\bibitem{HybridSort}
M.~Yang, G.~Han, B.~Yan, W.~Zhang, J.~Qi, H.~Lu, and D.~Wang, ``Hybrid-sort: Weak cues matter for online multi-object tracking,'' in \emph{Proceedings of the AAAI conference on artificial intelligence}, vol.~38, no.~7, 2024, pp. 6504--6512.

\bibitem{OCSort}
J.~Cao, J.~Pang, X.~Weng, R.~Khirodkar, and K.~Kitani, ``Observation-centric sort: Rethinking sort for robust multi-object tracking,'' in \emph{Proceedings of the IEEE/CVF conference on computer vision and pattern recognition}, 2023, pp. 9686--9696.

\bibitem{Adaptrack}
K.~Shim, K.~Ko, J.~Hwang, and C.~Kim, ``Adaptrack: Adaptive thresholding-based matching for multi-object tracking,'' in \emph{2024 IEEE International Conference on Image Processing (ICIP)}.\hskip 1em plus 0.5em minus 0.4em\relax IEEE, 2024, pp. 2222--2228.

\bibitem{CMTrack}
K.~Shim, J.~Hwang, K.~Ko, and C.~Kim, ``A confidence-aware matching strategy for generalized multi-object tracking,'' in \emph{2024 IEEE international conference on image processing (ICIP)}.\hskip 1em plus 0.5em minus 0.4em\relax IEEE, 2024, pp. 4042--4048.

\bibitem{DeepOCsort}
G.~Maggiolino, A.~Ahmad, J.~Cao, and K.~Kitani, ``Deep oc-sort: Multi-pedestrian tracking by adaptive re-identification,'' in \emph{2023 IEEE International conference on image processing (ICIP)}.\hskip 1em plus 0.5em minus 0.4em\relax IEEE, 2023, pp. 3025--3029.

\end{thebibliography}
\end{document}